\documentclass[10pt]{article}

\usepackage[a4paper,textwidth=18cm,textheight=24cm,top=2.85cm, bottom=2.85cm, left=1.5cm, right=1.5cm]{geometry}

\usepackage{template/icdp2009}

\makeatletter
\long\def\@makecaption#1#2{%
\vskip\abovecaptionskip
\sbox\@tempboxa{#1. #2}%
\ifdim \wd\@tempboxa >\hsize
#1. #2\par
\else
\global \@minipagefalse
\hb@xt@\hsize{\box\@tempboxa\hfil}%
\fi
\vskip\belowcaptionskip}
\makeatother

\usepackage{lmodern}

\usepackage{graphicx}
\usepackage{times}

\usepackage[numbers]{natbib}

\usepackage{amsmath,amsfonts}
\usepackage{algorithmic}
\usepackage{algorithm}
\usepackage{array}
\usepackage[caption=false,font=normalsize,labelfont=sf,textfont=sf]{subfig}
\usepackage{textcomp}
\usepackage{stfloats}
\usepackage{url}
\usepackage{verbatim}
\usepackage{graphicx}

\usepackage{hyperref}
\usepackage{cleveref}
\usepackage{dsfont}
\usepackage{tabularx}
\usepackage{booktabs}
\usepackage{makecell}
\usepackage{multirow}
\usepackage[justification=centering]{caption}
\usepackage{soul}
\usepackage[dvipsnames]{xcolor}
\usepackage{newfloat}
\usepackage{listings}
\DeclareCaptionStyle{ruled}{labelfont=normalfont,labelsep=colon,strut=off}
\floatstyle{ruled}
\newfloat{listing}{tb}{lst}{}
\floatname{listing}{Listing}

\begin{document}
\noindent

\title{Align and Adapt: Multimodal Multiview Human Activity Recognition under Arbitrary View Combinations}

\authorname{Duc-Anh Nguyen$^\#$, Nhien-An Le-Khac$^+$}
\authoraddr{
$^\#$ duc-anh.nguyen@ucdconnect.ie, nda97531@proton.me\\
$^+$ an.lekhac@ucd.ie\\
$^{\#+}$ University College Dublin, Ireland
}


\maketitle


\abstract
Multimodal multiview learning seeks to integrate information from diverse sources to enhance task performance. Existing approaches often struggle with flexible view configurations, including arbitrary view combinations, numbers of views, and heterogeneous modalities. Focusing on the context of human activity recognition, we propose AliAd, a model that combines multiview contrastive learning with a mixture-of-experts module to support arbitrary view availability during both training and inference. Instead of trying to reconstruct missing views, an adjusted center contrastive loss is used for self-supervised representation learning and view alignment, mitigating the impact of missing views on multiview fusion. This loss formulation allows for the integration of view weights to account for view quality. Additionally, it reduces computational complexity from $O(V^2)$ to $O(V)$, where $V$ is the number of views. To address residual discrepancies not captured by contrastive learning, we employ a mixture-of-experts module with a specialized load balancing strategy, tasked with adapting to arbitrary view combinations. We highlight the geometric relationship among components in our model and how they combine well in the latent space. AliAd is validated on four datasets encompassing inertial and human pose modalities, with the number of views ranging from three to nine, demonstrating its performance and flexibility.

\section{Introduction}
\label{sec: intro}
In multimodal multiview human activity recognition (HAR), parallel data sequences are recorded by sensor units (multiview) of the same or different sensor types (multimodal). The data format depends on sensor types. For instance, at each time step, an accelerometer records a 3D vector $[x,y,z]$, while a camera records a frame or a human pose.
Since each sensor has a unique perspective, which is a decisive factor for HAR accuracy \cite{Nguyen2024sok}, sensor fusion can provide more information and enhance accuracy. Early studies on multimodal multiview fusion have proposed to fuse views at the data, feature, or decision level \cite{Oh2024}, via concatenation or averaging \cite{Aguileta2019}.

Multimodal multiview systems often encounter the view-missing problem, which may arise from device or network failures. Also, sensors may be intentionally omitted during deployment to reduce costs. Relying on any fixed view combination can degrade the task performance when the available views do not align with the system's original design.

To handle missing views, existing methods often rely on missing-view indicators \cite{han2024fusemoe,yun2024flexmoe} or reconstruct the missing views using the available ones \cite{Woo2023,Wang_2023_CVPR,Liu2024mif}. However, such reconstruction inherently derives from the mutual information among views. In other words, they fill the missing views with information present in the observed views without recovering the actual missing information. Furthermore, training a separate reconstruction model for each view becomes impractical as the number of views increases. On the other hand, studies suggest that properly aligning modalities and ensuring coherence among their information can lead to more comprehensive and robust fusion \cite{Li2024}. Multiview contrastive learning pulls different views of the same data sample closer together in the latent space \cite{Tian2020}. It is also shown that contrastive learning has a distributional alignment effect \cite{Chen2024yourcl}. Recent work has integrated contrastive learning into multimodal multiview HAR \cite{Chen2024}. Since all sensors observe the same event, they share common information, making this scenario well-suited for contrastive learning, which leverages mutual information to extract robust features across views \cite{Bachman2019}.

In multimodal multiview data, some views may be more informative, while others may contain noise and irrelevant information. Using contrastive learning with these views can degrade high-quality views \cite{xu2023}. For instance, in cycling activity, an accelerometer on the leg is more indicative of the activity than one on the wrist. However, many contrastive learning methods neglect this and treat all views equally. Also, in multiview contrastive learning, the loss function is often computed between every view pair \cite{Tian2020}, resulting in an $O(V^2)$ time complexity where $V$ is the number of views.

To address the above problems, we propose AliAd (\textbf{Ali}gn and \textbf{Ad}apt), which supports arbitrary view combinations during both training and inference.
Each view is first processed by a feature extractor, which can be dedicated to that view or shared among homogeneous views. The resulting features are then combined using an attention-based weighting mechanism. A contrastive loss is used for self-supervised representation learning and to align views, thereby mitigating the impact of missing views on the fusion. This contrastive loss function also takes view quality into account. The model head responsible for the main task employs a sparse mixture-of-experts \cite{shazeer2017} architecture to address the remaining discrepancies among different view combinations. The contributions of this paper are summarized as follows:
\begin{itemize}
\item We propose AliAd, a multimodal multiview model capable of handling missing views during both training and inference. It shows that view alignment, without missing view reconstruction or filling, can robustly tackle the view missing problem.

\item Our adjusted center contrastive loss mitigates the impact of view missing on the fusion by aligning views in the hyperspherical latent space. It takes view quality into account and reduces time complexity.

\item A mixture-of-experts module equipped with a specialized load balancing strategy addresses the discrepancies left among different view combinations and generalizes to unseen view combinations.

\item We highlight the geometric properties of components in our model, and how they combine well together in a hyperspherical latent space.

\item Strong empirical results demonstrate the effectiveness of the proposed method and its robustness to missing views compared to baseline methods.
\end{itemize}

\section{Related Work}
\label{sec: related work}
\subsection{Contrastive Learning}
Multiview contrastive learning has proven to be an effective self-supervised representation learning tool. When computing contrastive loss with more than two views, full graph is a typical approach where the loss is computed for all view pairs \cite{Bian2023,Xu2022,Roy2021}. In contrast, the core view approach \cite{Tian2020} contrasts a core view with other views. It has been shown that full graph performs better as it does not rely on a single core view \cite{Lin2023}. For $V$ views, full graph involves $V(V{-}1)/2$ pairs (i.e., time complexity $O(V^2)$), which increases training time when the number of views is large. COCOA \cite{Deldari2022} modifies the positive and negative sampling strategy to reduce the number of pairs, thereby lowering time complexity. Alternatively, \cite{Ke2021,Wu2024} contrast each view with a concatenation-based view fusion; however, this approach is not flexible to missing views. \cite{Zhu2023} contrasts individual views with their summation-based fusion, but their adaptive fusion mechanism is restricted to the case of two views and graph-structured data.

To prevent the degradation of high-quality views while contrasting with low-quality ones, \cite{xu2023} assigns a weight to each pair in the full graph, encouraging more related views to be aligned more strongly. \cite{Jain2022} uses maximum mean discrepancy between views to select pairs, encouraging close positives and hard negatives. These methods compute pairwise weights from the feature distributions using discrepancy metrics. 
In our method, we train the contrastive loss and the main task in a joint learning setting; thus, view weights can be learned from the main task using an attention module.

\subsection{Mixture of Experts (MoE)}
Recently, sparse MoE \cite{shazeer2017} has been gaining attention from researchers, especially in the fields of natural language processing and computer vision. Each sparse MoE layer contains a set of sub-networks (experts), and for each input, a gating function activates a subset of specialized experts, enabling conditional computation. Most studies employ it within the Transformer architecture to reduce computation while preserving model capacity, facilitating model scaling \cite{fedus2022,Cai_2025}. MoE has been used with contrastive learning for feature learning \cite{mustafa2022} and stabilizing MoE's gating function \cite{Luo2024}. Some studies have applied MoE to multimodal and multiview learning tasks beyond the language-vision domain. For example, it has been used in multiview clustering \cite{Zhang2025}, brain tumor detection \cite{Liu2024}, Alzheimer’s disease tracking \cite{yun2024flexmoe}, sentiment analysis, and more \cite{han2024fusemoe}. Notably, the last two papers leverage MoE in Transformer layers to handle missing modalities, where each expert is specialized in different modality combinations.

\subsection{Multimodal Multiview Learning}
Missing modality is a common challenge in multimodal multiview systems. In some cases, views may be intentionally omitted during inference to reduce costs. Several studies have addressed this by leveraging multiple views during training and employing a fixed subset of views for inference, using approaches such as co-training with missing modalities \cite{Rahate2022} and contrastive learning \cite{Jain2022,Nguyen2024virtual}. To offer greater generality and reduce assumptions about available views, many works address arbitrary missing modalities. For example, the model can be trained to reconstruct missing modalities from the available ones \cite{Luo2023,Geraghty2025}. Some studies employ modality alignment techniques; however, these methods still require missing modality reconstruction afterwards \cite{Park2023,Wu2025}. Some studies fill missing data with trainable embeddings, acting as missing data indicators \cite{yun2024flexmoe,han2024fusemoe}. Another line of work aligns latent distributions among modalities, then imputes missing modalities with the average of the available ones \cite{Wang_2023_CVPR}. However, many methods lack full flexibility: some are limited to homogeneous views, others require particular view combinations in the training set.

\section{Proposed Method}
\begin{figure*}
  \centering
  \includegraphics[width=0.76\textwidth]{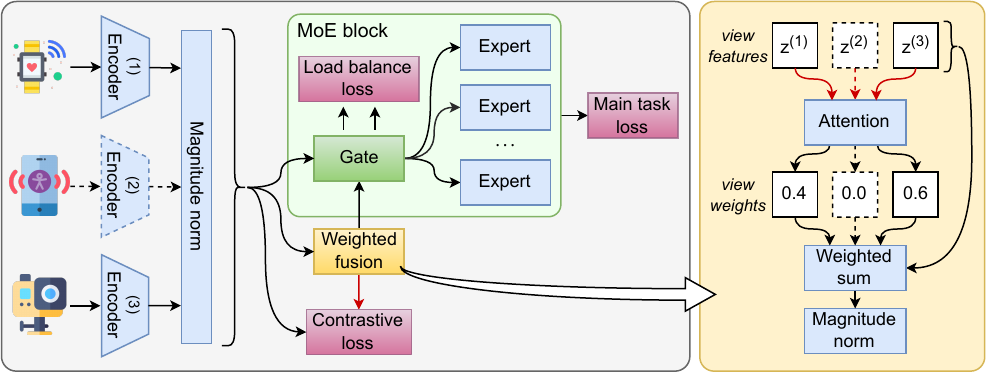}
  \caption{Overview of AliAd. Dashed lines indicate a missing view, which is excluded from computation. Red arrows represent stop-gradient connections, while black arrows allow gradient flow during backpropagation. During inference, the \textit{Weighted fusion} block's output is the only input to the \textit{MoE} block.}
  \label{fig: main}
\end{figure*}

Suppose we have a training set $\{(x_i^{(v)},y_i)|i=1,...,N;v=1,...,V\}$, where $N$ is the number of data samples, $V$ is the number of views. The views may correspond to the same or different modalities. The training set may contain missing views and missing labels. We train a model that can operate on any given view combination:
\begin{equation}\begin{aligned}
f_\theta^{(v)}&:x^{(v)} \to z^{(v)}; v \in \mathcal{S} \\
c_\theta&:z^{(\mathcal{S})} \to \hat{y}
\end{aligned}\end{equation}
where $f_\theta$ is the encoder, $z$ is the latent vector, $\mathcal{S}$ denotes the set of available views, and $c_\theta$ is the model head outputting class logits $\hat{y}$.

\Cref{fig: main} illustrates the AliAd model. The input data from each view is passed through its corresponding encoder to extract feature vectors, which are then normalized to the same magnitude. An attention module assigns a weight to each view within a data sample based on task relevance, enabling view fusion via a weighted sum. The fused vector also goes through magnitude normalization, putting it onto the same hypersphere as the individual views. During training, both individual and fused representations are passed into the MoE head for classification and are also used for contrastive learning. This ensures the model learns both useful view-specific information and mutual information among views.

All samples, including the unlabeled ones, contribute to the contrastive loss, while only labeled samples train the classifier. At inference time, only the fused representation is input to the classifier. In the following sections, we provide explanations and justifications for the components of the proposed model.

\subsection{Adjusted Center Contrastive Loss}\label{sec: accl}
\subsubsection{Why Contrastive Loss?}
Existing methods often handle missing views by using missing data indicators, which can be learnable embeddings \cite{han2024fusemoe,yun2024flexmoe} or zero/random values \cite{wu2024survey}. This approach informs the model which data samples have missing views by filling those missing entries with indicators, but it does not provide any additional information beyond indicating absence.
Many other methods reconstruct the missing views using the available ones \cite{Luo2023,Park2023,Wu2025,Geraghty2025}. Although this approach seems to recover the missing data, the reconstruction fundamentally depends on the information already present in the observed views. In other words, the reconstructed data inherently derives from the mutual information among views. Furthermore, training a reconstruction model for each view is less practical when the number of views gets large.

From the information theory point of view, when we train a model to reconstruct a missing view $C$ from the observed views $A$ and $B$, we are trying to minimize the conditional entropy:
\begin{equation}
H(C|A,B)=H(C) - I(C;A,B)
\end{equation}
where $H(C)=-\int p(c) \log p(c) \, dc$ is the intrinsic entropy of $C$ determined by its marginal distribution $p(c)$, and is not changed by the reconstruction model. Therefore, the model is essentially learning to maximize the mutual information $I(C; A, B)$. If we directly train the model to maximize this mutual information, it can achieve the same effect without constructing redundant data, thereby reducing workload.
Prior studies have shown that multiview contrastive learning can maximize mutual information among views \cite{Bachman2019,Tian2020}, making it suitable for this task.

\begin{figure}[]
  \centering
  \includegraphics[width=\linewidth]{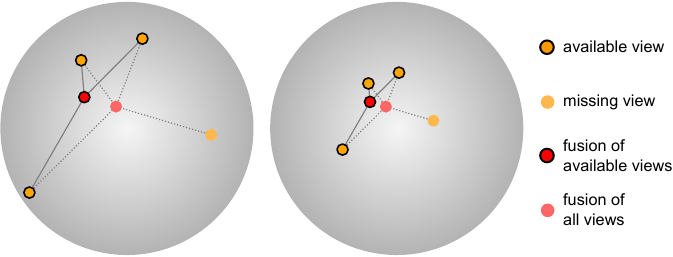}
  \caption{Illustration of fusion robustness to view missing. When views are closer together (right), the fusion shifts less upon view removal than when views are dispersed (left).}
\label{fig: hypersphere fusion}
\end{figure}

We also use contrastive loss for two other reasons.
First, contrastive loss pulls similar data points (positives) together and pushes dissimilar ones (negatives) apart \cite{wang2020sphere}. In this multiview setting, we consider views of the same data sample as positives and views from different samples in a training batch as negatives. Since we fuse views by finding their center in the feature space, the number and distribution of views influence the fusion. By pulling views closer together, contrastive loss improves the fusion's robustness to view missing (\Cref{fig: hypersphere fusion}).

Second, contrastive learning is widely used for self-supervised representation learning \cite{Liu2023,Chen2024}. Even without fusion, studies have shown that multiview contrastive learning can improve uniview models \cite{Jain2022,Nguyen2024virtual}.

\subsubsection{Loss Function Definition}
\begin{figure}[]
  \centering
  \includegraphics[width=0.65\linewidth]{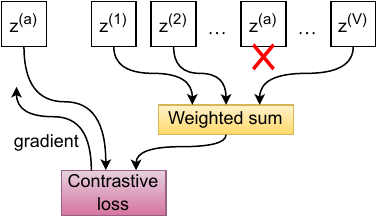}
  \caption{Adjusted center contrastive loss. Each view is contrasted with the other views' center on the hypersphere.}
  \label{fig: wfcl}
\end{figure}

For each data sample at index $i$, the contrastive loss $\mathcal{L}_{\text{pair}}$ between two views $a$ and $b$ is:
\begin{align}
&\ell^{(a,b)} = -\log\frac{g(z^{(a)}_i,z^{(b)}_i)}{\sum\limits_{j}\sum\limits_{v\in \{a,b\}} \mathds{1}_{[j\ne i\ OR\ v\ne a]} g(z^{(a)}_i,z^{(v)}_j)} \nonumber \\[5pt]
&\mathcal{L}_{\text{pair}}(z^{(a)},z^{(b)})=\ell^{(a,b)}+\ell^{(b,a)}
\label{eq: lpair}
\end{align}
where $i$ and $j$ are sample indices within a batch. The function $g(\cdot)$ computes the exponentiated cosine similarity, scaled by a temperature hyperparameter $\tau$:
\begin{equation}
\label{eq: cosine critic}
g(z^{(a)},z^{(b)}) = \exp \left(\frac{z^{(a)}\cdot z^{(b)}}{\|z^{(a)}\|\cdot\|z^{(b)}\|} \cdot \frac{1}{\tau} \right)
\end{equation}
To compute multiview contrastive loss, the full graph approach contrasts all pairs, indirectly pulling all views together. The core view approach contrasts each view with a designated core view, aligning all views toward this core \cite{Tian2020}. Full graph often yields better results than core view, as it does not rely on a single view \cite{Lin2023}. However, core view is more efficient, requiring only $V{-}1$ pairs compared to $V(V{-}1)/2$ pairs in a full graph. Our method instead aligns all views directly to the center by contrasting each view with the center of the other views (\Cref{fig: wfcl}). This reduces time complexity while preserving the benefits of the full graph approach. Also, view weights can be integrated straightforwardly to adjust the center, enabling control over the influence of individual views with varying quality. Conversely, a full graph approach would require assigning a pairwise weight to every view pair.

As the objective is to pull each view closer to the center, we treat the center as a constant, allowing gradients to flow only through the individual views. Since cosine similarity is used in contrastive loss, vector magnitude can be ignored, and a simple summation places the result vector at the angular center of the constituent vectors. Specifically, our adjusted center contrastive loss is:
\begin{equation}
\label{eq: wfcl}
\mathcal{L}_{\text{AC}}=\frac{1}{V{-}1} \sum\limits_{a}^V \underbrace{(1-w^{(a)})}_{\text{stop grad}}\mathcal{L}_{\text{pair}} \biggl( z^{(a)}, \underbrace{\sum\limits_{v\ne a}^V w^{(v)}z^{(v)}}_{\text{stop grad}} \biggl)
\end{equation}
The weight term $w(v)$ puts the center closer to higher-quality views, causing all views to be pulled more toward those of higher quality. The term $1{-}w^{(a)}$ acts as a loss weight, assigning a lower weight when the loss function tries to pull a high-quality view toward others. This weight will be discussed in \Cref{sec: view attn module}. The loss term is divided by $V{-}1$ instead of $V$ to compensate for the scale decrease caused by the weight $1{-}w^{(a)}$. Specifically, the total scaling factor is:
\begin{equation}
\frac{\sum_a^V(1-w^{(a)})}{V-1} = \frac{V-\sum_a^V w^{(a)}}{V-1} =\frac{V-1}{V-1} = 1
\end{equation}
So \Cref{eq: wfcl} is a weighted average of $\mathcal{L}_{\text{pair}}$ across views.

\subsubsection{Loss Function Implementation}\label{sec: accl impl}
\begin{listing}[]
\caption{Python-style pseudocode for adjusted center contrastive loss}\label{lst: wfcl}
\begin{lstlisting}[
    language=Python,
    morekeywords={stop_gradient,contrast_pair,sg},
    basicstyle=\fontsize{9.2}{10}\selectfont\ttfamily
]
# z: feature vectors of all views [VxNxC]
# w: attention weights of all views [VxN]

num_views, batch, channels = z.shape
w = stop_gradient(w)
wz = stop_gradient(z * w)
center = sum(wz, dimension=0)  #[NxC]
L = 0
for i in range(num_views):
    zi = z[i]  #[NxC]
    c = center - wz[i]
    L += contrast_pair(zi, c) * (1 - w[i])
return L / (num_views - 1)
\end{lstlisting}
\end{listing}

\Cref{lst: wfcl} demonstrates how $\mathcal{L}_\text{AC}$ in \Cref{eq: wfcl} is implemented. The function \textit{contrast\_pair} is the contrastive loss between a pair of views $\mathcal{L}_\text{pair}$ defined in \Cref{eq: lpair}. In this implementation, we compute $\sum_{v}^V w^{(v)} z^{(v)} {-} w^{(a)} z^{(a)}$ instead of $\sum_{v\ne a}^V w^{(v)} z^{(v)}$ as in \Cref{eq: wfcl}. Although they produce identical results, the former avoids re-computing the fusion every iteration, i.e., $O(V^2N)$. It computes the fusion once and subtracts each individual view from that fusion in the loop, i.e., $O(VN{+}VN)$. Including the \textit{contrast\_pair} function, the total time complexity of \Cref{lst: wfcl} is $O(VN {+} V(N^2 {+} N))$, which simplifies to $O(VN^2)$.

The full graph approach, which considers all possible view pairs, has a complexity of $O(V^2N^2)$. COCOA \cite{Deldari2022} reduces this by removing cross-view negative pairs, resulting in $O(V^2N {+} VN^2)$. In contrast, our method combines views rather than removing them, achieving $O(VN^2)$. Any multiview contrastive loss function has a minimum time complexity of $O(N^2)$, as this is required for computing a single view pair. Assuming that the number of data samples $N$ is fixed across methods, we focus on the dependency on $V$. Without considering $N$, our approach improves the time complexity from $O(V^2)$ to $O(V)$.

\subsection{Attention-based View Fusion}\label{sec: view attn module}
\begin{figure}[]
\centering
\subfloat[Average fusion]{\includegraphics[width=0.48\linewidth]{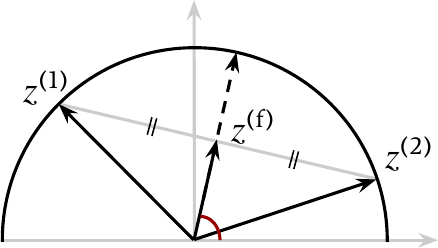}}\hfill
\subfloat[Weighted fusion]{\includegraphics[width=0.48\linewidth]{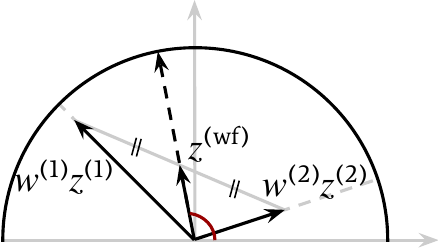}}
\caption{Fusion of two vectors. The fused vector $z^{(wf)}$ is oriented more toward the vector with a higher weight.}\label{fig: fusion circles}
\end{figure}
We employ an attention module to capture differences in data quality and task relevance among views. Fusion is performed via a weighted sum followed by vector magnitude normalization. As a result, the fusion lies in the center of the individual views, while being closer to high-quality views on the hypersphere (\Cref{fig: fusion circles}). This not only improves the fusion for the main task, but also alleviates the degradation of high-quality views when contrasted with low-quality ones using the adjusted center contrastive loss.

Following prior work \cite{Yuan2018,Ma2019,Yang2020}, we process each view independently without explicitly modeling cross-view interactions. This design naturally accommodates missing views without missing data indicators or reconstruction. Also, because it processes views separately, it is not affected by unseen view combinations.

Specifically, we use a shared MLP network containing two fully connected layers and a ReLU activation in between to compute a scalar weight for each view, and a softmax function to normalize view weights. We stop the backward gradient flow between the encoders and the attention module (\Cref{fig: main}) to ensure only the attention module learns view importance, while the encoders focus exclusively on feature extraction. View weights are computed as follows:
\begin{equation}\label{eq: attn weight}
w^{(v)} = \text{softmax}(\text{MLP}(\underbrace{z\vphantom{|}}_{\text{stop grad}}))_v
\end{equation}
Then, the weighted fusion is:
\begin{equation}
z^{(wf)} = \sum_v^V w^{(v)}z^{(v)}
\end{equation}
This fused representation is used for both classification and contrastive loss. Since the view attention module is trained with the main task’s loss function, it learns view importance specific to the main task.

\subsection{Mixture of Expert Classifiers}
We use a MoE for the classification head to further process residual discrepancies among view combinations not captured by the contrastive loss. Each expert in the MoE specializes in handling different patterns, and the gating mechanism routes inputs to the most suitable experts.

The previous sections demonstrate that individual views are pivotal in the fusion process, especially with missing views. Therefore, we train the classification head using both the fused representation and each individual view. While contrastive learning captures mutual information across views, training on individual views extracts useful view-specific features. Also, since the fusion lies at the center of the individual views, this training approach strengthens the robustness to arbitrary, unseen view combinations.

Because the fusion and individual views are trained for different purposes, we design a load balancing strategy that separates them, preventing the individual views from occupying part of the expert pool.

\subsubsection{MoE Head}
We adopt the sparse MoE architecture with a noisy top-K gating network \cite{shazeer2017}, where each input token is processed by K experts. Other studies often use MoE in conjunction with the Transformer architecture \cite{yun2024flexmoe,han2024fusemoe}, modeling cross-view interaction. Because our model processes views separately, we integrate MoE without the self-attention layer commonly seen in Transformer. Each expert is an MLP network outputting class logits. The MoE classification head is:
\begin{equation}
\hat{y} = \sum_{e}^E \text{Gate}(z, K)_e \text{Expert}_e(z)
\end{equation}
For each input $z$ of an individual view or the fusion, the gate outputs a set of weights across the $E$ experts. Only the top $K$ experts are kept, and their weights are passed into a softmax function, while the rest are assigned a weight of 0.

Finally, since our main task is classification, the cross-entropy loss function $H$ is used:
\begin{equation}
\mathcal{L}_\text{cls} = \frac{1}{V+1}\left( H(y,\hat{y}^{(wf)}) + \sum_v^V H(y,\hat{y}^{(v)}) \right)
\end{equation}

\subsubsection{Load Balancing}
The load balancing loss encourages the gate to distribute input tokens more evenly among the experts, preventing expert overuse or underuse. As we train individual views and the fusion together, combinations with only one view are trained much more than any other combinations that may appear in the fusion. Consequently, the whole set of tokens is encouraged to spread evenly, but one-view and multi-view combinations may be allocated to disjoint sets of experts. If this happens, it will nullify the purpose of training individual views to strengthen the model's robustness to unseen view combinations.

To address this, we compute the load balancing loss below for the individual views and the fusion separately, promoting both to spread evenly across experts.
\begin{equation}
\mathcal{L}_\text{LB} = CV^2(\{importance_e\}_e^E)+CV^2(\{load_e\}_e^E)
\end{equation}
The \textit{importance} for each expert quantifies the total gating weight assigned to that expert across tokens; the \textit{load} measures how many tokens are dispatched to each expert, indicating the actual token count; $CV^2$ is the coefficient of variation squared function $CV^2(x) = (\sigma(x)/\mu(x))^2$.

\subsection{AliAd in A Hyperspherical Feature Space}\label{sec: attn wfcl synergy}
In this section, we look at the geometric properties of AliAd's components and their compatibility in a hyperspherical latent space.

In contrastive learning, cosine similarity is the most common similarity measure \cite{Chen2024}. This approach is often viewed as learning features within a hyperspherical space \cite{wang2020sphere}. By using cosine similarity, contrastive learning encodes information in the angular relationships among vectors while ignoring their magnitudes.

Attention-based view fusion also works well in a hyperspherical space. Views are fused using a weighted sum. When a view weight is applied to a vector, it changes the magnitude of this vector. The fusion will be shifted closer in angle to the longer constituent vectors. If feature vectors are placed into a hyperspherical space, the attention module does not need to account for differences in initial magnitudes. \Cref{fig: fusion circles} illustrates how two views in a hyperspherical latent space are fused without and with weights; the same principle applies to fusion of more than two views.

We train the MoE classification head using the fused view and all individual views. Since any combination of views resides within the hyperspherical convex hull defined by the individual views (\Cref{fig: hypersphere fusion}), this training strategy equips the gate and the expert models to be more robust to arbitrary, unseen view combinations.

Although the model can still operate with features in a Euclidean space, we project all representations onto a hyperspherical space to ensure consistency across its components. The magnitude normalization below places feature vectors across all views onto a hypersphere. The square root of the vector dimension ensures the feature scale is independent of dimension, preventing excessively small scales for high-dimensional vectors.
\begin{equation}
\text{MagNorm}(z) = \frac{z}{\|z\|} \cdot \sqrt{\text{dim}(z)}
\end{equation}

\section{Experiment}

\subsection{Experimental Setup}

\subsubsection{Datasets}
\begin{table}[]
\centering
\fontsize{9}{10}\selectfont
\setlength{\tabcolsep}{1.5mm} 
\begin{tabular}{@{}lllr@{}}
\toprule
Dataset & \makecell[l]{No. subjects\\train/ valid/ test} & No. views & \makecell[r]{No.\\classes} \\ \midrule
CMDFall & 25 / 5 / 20 & accel.$\times2$; 3D pose$\times5$ & 20 \\
Daily Sport & 3 / 2 / 3 & accel.$\times5$ & 19 \\
RealDisp & 7 / 3 / 7 & accel.$\times9$ & 33 \\
UP-Fall & 7 / 3 / 7 & accel.$\times4$; 2D pose$\times1$ & 11 \\ \bottomrule
\end{tabular}
\caption{Datasets used in experiments.}
\label{tbl: datasets}
\end{table}
We use four datasets: CMDFall \cite{tran2018}, Daily and Sport Activities \cite{ALTUN2010}, UP-Fall \cite{martinez2019}, and RealDisp \cite{banos2012}. They were chosen for data quantity and diversity, number of views, and annotation granularity. For example, in CMDFall, activities are performed continuously, promoting natural and diverse movements; RealDisp has many sensors and classes. All datasets comprise complex human activities, providing greater discriminative power for robust model evaluation.

Each dataset is divided by subject IDs into training, validation, and test sets. \Cref{tbl: datasets} summarizes dataset information. For fairness, we use the same labeled training set for all methods in the comparison, including methods using self-supervised contrastive loss. F1-score is used as the evaluation metric to address class imbalance in the test sets.

We use sliding window to create data samples from the raw time series. For all datasets except DailySport, we use a window size of 4 seconds, which is sufficient to capture activities in the label lists. For the DailySport dataset, raw data are originally formatted as 5-second windows.

\subsubsection{Baselines}
The following methods are included in the comparison: CMC \cite{Tian2020}, COCOA \cite{Deldari2022}, Flex-MoE \cite{yun2024flexmoe}, FuseMoE \cite{han2024fusemoe}, and ShaSpec \cite{Wang_2023_CVPR}. CMC and COCOA are not explicitly designed for missing views, but since they use contrastive learning and process views independently, missing views are naturally handled the same way as in our method (\Cref{fig: hypersphere fusion}). Flex-MoE and FuseMoE both use learnable embeddings as missing view indicators and employ MoE to accommodate arbitrary view combinations. They also use the transformer architecture to model cross-view interactions. ShaSpec learns separate representations for view-specific and shared information; any missing view is reconstructed from the shared features of the available views.

\subsubsection{Configurations}
\begin{table}[]
\centering
\fontsize{10}{10}\selectfont
\begin{tabular}{@{}lr@{}}
\toprule
Hyperparameter & Value \\
\midrule
optimizer & Adam \\
learning rate & 10$^{-3}$ \\
classification batch size & 16 \\
contrastive learning batch size & 16 \\
classification loss weight & 1 \\
contrastive loss weight & 1 \\
load balancing loss weight & 10$^{-2}$ \\
temperature $\tau$ in contrastive loss & 0.1 \\
number of experts & 16 \\
top K experts & 3 \\
\bottomrule
\end{tabular}
\caption{Hyperparameters for AliAd}
\label{tbl: hyperparam}
\end{table}

For all experiments and methods, we implement a lightweight 1D CNN based on ResNet \cite{He_2016_CVPR} with 4 residual blocks as the encoder network. Scaling and time warping augmentation techniques \cite{um2017} are applied to training data. Additionally, 3D rotation is used for accelerometer data, rotation around the Z-axis is applied to 3D poses, and horizontal flipping is applied to 2D poses. Every model is trained for at least 20 epochs, and the best model checkpoint, determined using the validation set, is evaluated on the test set. The batch size is tuned between 8, 16, and 32, and the learning rate is tuned between $10^{-3}$, $10^{-4}$, and $10^{-5}$.
For the proposed method, we tune the number of experts between 8, 16, and 32, and top K between 2, 3, and 4. The final hyperparameters of our method are shown in \Cref{tbl: hyperparam}.
All reported scores are averages of three runs with three fixed random seeds.

\subsection{Experimental Results}

\subsubsection{Complete Training Data}
\begin{table*}[]
\fontsize{8}{10}\selectfont
\setlength{\tabcolsep}{1mm} 
\centering
\begin{tabularx}{\textwidth}{@{}
l
*{12}{>{\centering\arraybackslash}X}
@{}}
\Xhline{2\arrayrulewidth}

Dataset & 
\multicolumn{3}{|c}{CMDFall} & 
\multicolumn{3}{|c}{DailySport} & 
\multicolumn{3}{|c}{RealDisp} & 
\multicolumn{3}{|c}{UP-Fall} \\

No. views & 
\multicolumn{1}{|>{\centering\arraybackslash}X}{1} & 
2 & 
3 & 
\multicolumn{1}{|>{\centering\arraybackslash}X}{1} & 
3 & 
5 & 
\multicolumn{1}{|>{\centering\arraybackslash}X}{1} & 
5 & 
9 & 
\multicolumn{1}{|>{\centering\arraybackslash}X}{1} & 
3 & 
5 \\

\hline

CMC$^{*}$ & 59.59{\fontsize{6.5}{0}\selectfont$\pm$0.2}&70.80{\fontsize{6.5}{0}\selectfont$\pm$0.6}&76.93{\fontsize{6.5}{0}\selectfont$\pm$0.9}&77.09{\fontsize{6.5}{0}\selectfont$\pm$1.3}&87.03{\fontsize{6.5}{0}\selectfont$\pm$1.5}&92.05{\fontsize{6.5}{0}\selectfont$\pm$1.6}&75.79{\fontsize{6.5}{0}\selectfont$\pm$1.2}&95.06{\fontsize{6.5}{0}\selectfont$\pm$0.4}&97.32{\fontsize{6.5}{0}\selectfont$\pm$0.1}&71.17{\fontsize{6.5}{0}\selectfont$\pm$0.6}&\hl{87.31{\fontsize{6.5}{0}\selectfont$\pm$0.1}}&91.63{\fontsize{6.5}{0}\selectfont$\pm$0.5} \\

COCOA$^{*}$ & 56.33{\fontsize{6.5}{0}\selectfont$\pm$0.6}&69.29{\fontsize{6.5}{0}\selectfont$\pm$0.4}&76.40{\fontsize{6.5}{0}\selectfont$\pm$0.6}&75.26{\fontsize{6.5}{0}\selectfont$\pm$0.6}&86.31{\fontsize{6.5}{0}\selectfont$\pm$1.0}&90.68{\fontsize{6.5}{0}\selectfont$\pm$1.4}&68.66{\fontsize{6.5}{0}\selectfont$\pm$0.5}&94.55{\fontsize{6.5}{0}\selectfont$\pm$0.2}&97.42{\fontsize{6.5}{0}\selectfont$\pm$0.2}&71.43{\fontsize{6.5}{0}\selectfont$\pm$0.5}&87.24{\fontsize{6.5}{0}\selectfont$\pm$0.8}&91.50{\fontsize{6.5}{0}\selectfont$\pm$1.6} \\

FlexMoE & 25.09{\fontsize{6.5}{0}\selectfont$\pm$3.8}&52.37{\fontsize{6.5}{0}\selectfont$\pm$2.3}&74.42{\fontsize{6.5}{0}\selectfont$\pm$0.5}&12.13{\fontsize{6.5}{0}\selectfont$\pm$2.3}&58.52{\fontsize{6.5}{0}\selectfont$\pm$4.6}&88.40{\fontsize{6.5}{0}\selectfont$\pm$1.4}&03.46{\fontsize{6.5}{0}\selectfont$\pm$1.6}&74.44{\fontsize{6.5}{0}\selectfont$\pm$2.0}&96.92{\fontsize{6.5}{0}\selectfont$\pm$0.3}&09.09{\fontsize{6.5}{0}\selectfont$\pm$1.3}&35.83{\fontsize{6.5}{0}\selectfont$\pm$4.5}&87.14{\fontsize{6.5}{0}\selectfont$\pm$1.1} \\

FuseMoE & 28.49{\fontsize{6.5}{0}\selectfont$\pm$1.1}&55.08{\fontsize{6.5}{0}\selectfont$\pm$0.6}&74.23{\fontsize{6.5}{0}\selectfont$\pm$1.0}&24.21{\fontsize{6.5}{0}\selectfont$\pm$2.7}&70.68{\fontsize{6.5}{0}\selectfont$\pm$2.1}&90.13{\fontsize{6.5}{0}\selectfont$\pm$0.8}&04.31{\fontsize{6.5}{0}\selectfont$\pm$1.4}&65.14{\fontsize{6.5}{0}\selectfont$\pm$5.7}&96.69{\fontsize{6.5}{0}\selectfont$\pm$0.5}&15.07{\fontsize{6.5}{0}\selectfont$\pm$0.6}&62.52{\fontsize{6.5}{0}\selectfont$\pm$0.9}&87.17{\fontsize{6.5}{0}\selectfont$\pm$0.5} \\

ShaSpec & 35.65{\fontsize{6.5}{0}\selectfont$\pm$0.1}&60.59{\fontsize{6.5}{0}\selectfont$\pm$0.7}&74.55{\fontsize{6.5}{0}\selectfont$\pm$0.5}&35.42{\fontsize{6.5}{0}\selectfont$\pm$1.6}&74.46{\fontsize{6.5}{0}\selectfont$\pm$0.7}&88.58{\fontsize{6.5}{0}\selectfont$\pm$1.5}&20.11{\fontsize{6.5}{0}\selectfont$\pm$4.7}&90.14{\fontsize{6.5}{0}\selectfont$\pm$1.4}&\hl{97.53{\fontsize{6.5}{0}\selectfont$\pm$0.5}}&47.01{\fontsize{6.5}{0}\selectfont$\pm$2.2}&75.62{\fontsize{6.5}{0}\selectfont$\pm$2.3}&90.88{\fontsize{6.5}{0}\selectfont$\pm$0.5} \\

AliAd & \hl{59.95{\fontsize{6.5}{0}\selectfont$\pm$0.4}}&\hl{71.21{\fontsize{6.5}{0}\selectfont$\pm$1.1}}&\hl{77.28{\fontsize{6.5}{0}\selectfont$\pm$1.2}}&\hl{80.56{\fontsize{6.5}{0}\selectfont$\pm$0.6}}&\hl{90.54{\fontsize{6.5}{0}\selectfont$\pm$0.7}}&\hl{93.64{\fontsize{6.5}{0}\selectfont$\pm$0.4}}&\hl{80.75{\fontsize{6.5}{0}\selectfont$\pm$0.1}}&\hl{95.55{\fontsize{6.5}{0}\selectfont$\pm$0.1}}&96.74{\fontsize{6.5}{0}\selectfont$\pm$0.2}&\hl{72.75{\fontsize{6.5}{0}\selectfont$\pm$0.5}}&86.29{\fontsize{6.5}{0}\selectfont$\pm$0.3}&\hl{92.17{\fontsize{6.5}{0}\selectfont$\pm$0.3}} \\

\Xhline{2\arrayrulewidth}
\end{tabularx}
\caption{F1-score (\%) comparison when the training set has complete views.}
\label{tbl: exp complete}
\end{table*}

We train the models on training sets with complete views and evaluate them on test sets with missing views to assess how they respond to unseen view combinations. For the CMDFall dataset, we use only 3 out of 7 views (1 skeleton and 2 accelerometer views), retaining only samples with all 3 views present. We do not use all 7 views as most samples in this dataset have missing views, hence it is not suitable for this experiment.
For each dataset, we evaluate the models under three scenarios: using a single view, half of the views, and all views. For each scenario, the test score is the average across all possible combinations of the specified number of views.

\Cref{tbl: exp complete} shows that our proposed method achieves the highest scores in 10 out of 12 tests. Methods that rely on missing view indicators (i.e., Flex-MoE and FuseMoE) exhibit reduced performance when evaluated on unseen view combinations, particularly in datasets with a large number of views, due to insufficient training of the indicator embeddings for those combinations.

\subsubsection{Missing Training Data}

\begin{table*}[]
\fontsize{8}{10}\selectfont
\setlength{\tabcolsep}{1mm} 
\centering
\begin{tabularx}{\textwidth}{@{}
l
*{12}{>{\centering\arraybackslash}X}
@{}}
\Xhline{2\arrayrulewidth}

Dataset & 
\multicolumn{3}{|c}{CMDFall} & 
\multicolumn{3}{|c}{DailySport} & 
\multicolumn{3}{|c}{RealDisp} & 
\multicolumn{3}{|c}{UP-Fall} \\

No. views & 
\multicolumn{1}{|>{\centering\arraybackslash}X}{1} & 
$\leq$ 4 &
$\leq$ 7 &
\multicolumn{1}{|>{\centering\arraybackslash}X}{1} & 
$\leq$ 3 & 
$\leq$ 5 & 
\multicolumn{1}{|>{\centering\arraybackslash}X}{1} & 
$\leq$ 5 & 
$\leq$ 9 & 
\multicolumn{1}{|>{\centering\arraybackslash}X}{1} & 
$\leq$ 3 & 
$\leq$ 5 \\

\hline

CMC$^*$ & 49.41{\fontsize{6.5}{0}\selectfont$\pm$0.7}&74.27{\fontsize{6.5}{0}\selectfont$\pm$0.3}&\hl{81.86{\fontsize{6.5}{0}\selectfont$\pm$0.4}}&74.24{\fontsize{6.5}{0}\selectfont$\pm$0.6}&82.72{\fontsize{6.5}{0}\selectfont$\pm$0.6}&87.69{\fontsize{6.5}{0}\selectfont$\pm$0.2}&68.86{\fontsize{6.5}{0}\selectfont$\pm$0.3}&84.80{\fontsize{6.5}{0}\selectfont$\pm$0.3}&91.55{\fontsize{6.5}{0}\selectfont$\pm$0.3}&70.84{\fontsize{6.5}{0}\selectfont$\pm$0.9}&82.45{\fontsize{6.5}{0}\selectfont$\pm$1.1}&87.44{\fontsize{6.5}{0}\selectfont$\pm$0.9} \\
COCOA$^*$ & 45.71{\fontsize{6.5}{0}\selectfont$\pm$0.7}&71.36{\fontsize{6.5}{0}\selectfont$\pm$0.5}&79.84{\fontsize{6.5}{0}\selectfont$\pm$0.3}&70.98{\fontsize{6.5}{0}\selectfont$\pm$0.1}&81.54{\fontsize{6.5}{0}\selectfont$\pm$0.3}&87.04{\fontsize{6.5}{0}\selectfont$\pm$0.5}&61.61{\fontsize{6.5}{0}\selectfont$\pm$0.7}&82.60{\fontsize{6.5}{0}\selectfont$\pm$0.4}&90.95{\fontsize{6.5}{0}\selectfont$\pm$0.1}&70.12{\fontsize{6.5}{0}\selectfont$\pm$0.6}&83.12{\fontsize{6.5}{0}\selectfont$\pm$0.7}&88.68{\fontsize{6.5}{0}\selectfont$\pm$1.1} \\
FlexMoE & 12.30{\fontsize{6.5}{0}\selectfont$\pm$0.9}&46.04{\fontsize{6.5}{0}\selectfont$\pm$1.2}&74.19{\fontsize{6.5}{0}\selectfont$\pm$0.6}&40.68{\fontsize{6.5}{0}\selectfont$\pm$2.9}&73.72{\fontsize{6.5}{0}\selectfont$\pm$0.9}&86.71{\fontsize{6.5}{0}\selectfont$\pm$0.9}&13.69{\fontsize{6.5}{0}\selectfont$\pm$0.6}&50.17{\fontsize{6.5}{0}\selectfont$\pm$0.6}&70.24{\fontsize{6.5}{0}\selectfont$\pm$0.6}&54.42{\fontsize{6.5}{0}\selectfont$\pm$2.4}&78.22{\fontsize{6.5}{0}\selectfont$\pm$0.4}&87.81{\fontsize{6.5}{0}\selectfont$\pm$0.5} \\
FuseMoE & 09.60{\fontsize{6.5}{0}\selectfont$\pm$0.7}&47.65{\fontsize{6.5}{0}\selectfont$\pm$1.2}&76.22{\fontsize{6.5}{0}\selectfont$\pm$0.9}&49.06{\fontsize{6.5}{0}\selectfont$\pm$2.6}&76.46{\fontsize{6.5}{0}\selectfont$\pm$1.9}&87.62{\fontsize{6.5}{0}\selectfont$\pm$2.1}&28.25{\fontsize{6.5}{0}\selectfont$\pm$0.9}&67.87{\fontsize{6.5}{0}\selectfont$\pm$0.3}&84.35{\fontsize{6.5}{0}\selectfont$\pm$0.5}&52.10{\fontsize{6.5}{0}\selectfont$\pm$1.8}&78.17{\fontsize{6.5}{0}\selectfont$\pm$0.5}&88.69{\fontsize{6.5}{0}\selectfont$\pm$0.8} \\
ShaSpec & 24.61{\fontsize{6.5}{0}\selectfont$\pm$1.1}&61.10{\fontsize{6.5}{0}\selectfont$\pm$0.5}&77.73{\fontsize{6.5}{0}\selectfont$\pm$0.1}&61.77{\fontsize{6.5}{0}\selectfont$\pm$0.5}&78.63{\fontsize{6.5}{0}\selectfont$\pm$0.5}&87.25{\fontsize{6.5}{0}\selectfont$\pm$0.3}&47.76{\fontsize{6.5}{0}\selectfont$\pm$2.1}&78.28{\fontsize{6.5}{0}\selectfont$\pm$1.2}&89.37{\fontsize{6.5}{0}\selectfont$\pm$0.9}&61.59{\fontsize{6.5}{0}\selectfont$\pm$3.3}&80.20{\fontsize{6.5}{0}\selectfont$\pm$1.8}&88.16{\fontsize{6.5}{0}\selectfont$\pm$1.1} \\
AliAd & \hl{50.29{\fontsize{6.5}{0}\selectfont$\pm$0.4}}&\hl{74.78{\fontsize{6.5}{0}\selectfont$\pm$0.6}}&81.75{\fontsize{6.5}{0}\selectfont$\pm$0.6}&\hl{79.68{\fontsize{6.5}{0}\selectfont$\pm$0.3}}&\hl{87.26{\fontsize{6.5}{0}\selectfont$\pm$0.5}}&\hl{91.25{\fontsize{6.5}{0}\selectfont$\pm$0.5}}&\hl{76.33{\fontsize{6.5}{0}\selectfont$\pm$0.1}}&\hl{88.18{\fontsize{6.5}{0}\selectfont$\pm$0.2}}&\hl{93.33{\fontsize{6.5}{0}\selectfont$\pm$0.1}}&\hl{74.98{\fontsize{6.5}{0}\selectfont$\pm$0.5}}&\hl{85.31{\fontsize{6.5}{0}\selectfont$\pm$0.5}}&\hl{89.72{\fontsize{6.5}{0}\selectfont$\pm$0.2}} \\

\Xhline{2\arrayrulewidth}
\end{tabularx}
\caption{F1-score (\%) comparison when the training set has missing views.}
\label{tbl: exp missing}
\end{table*}

To simulate missing views, we randomly drop each view in every sample with a probability of $\sqrt[V]{10^{-3}}$, where $V$ is the number of views, thus each sample has a 0.1\% chance of having all views dropped. Except for CMDFall, we use all 7 views without dropping data.
The evaluation scenarios are the same as in the previous experiment. Due to missing views, the specified numbers of views now represent upper bounds instead of fixed quantities.

\Cref{tbl: exp missing} shows that AliAd achieves the highest scores overall. The performance gap between AliAd and other methods becomes more pronounced, particularly on the DailySport and RealDisp datasets. This demonstrates AliAd's robustness to missing views compared to the baselines.

\subsection{Ablation Study}
\begin{table*}
\fontsize{8}{10}\selectfont
\setlength{\tabcolsep}{1mm} 
\centering
\begin{tabularx}{\linewidth}{@{}
l
*{1}{|>{\centering\arraybackslash}X}
*{11}{>{\centering\arraybackslash}X}
@{}}
\Xhline{2\arrayrulewidth}

Dataset & 
\multicolumn{3}{c}{CMDFall} & 
\multicolumn{3}{|c}{DailySport} & 
\multicolumn{3}{|c}{RealDisp} & 
\multicolumn{3}{|c}{UP-Fall} \\

No. views & 
\multicolumn{1}{>{\centering\arraybackslash}X}{1} & 
$\leq$ 4 &
$\leq$ 7 &
\multicolumn{1}{|>{\centering\arraybackslash}X}{1} & 
$\leq$ 3 & 
$\leq$ 5 & 
\multicolumn{1}{|>{\centering\arraybackslash}X}{1} & 
$\leq$ 5 & 
$\leq$ 9 & 
\multicolumn{1}{|>{\centering\arraybackslash}X}{1} & 
$\leq$ 3 & 
$\leq$ 5 \\

\hline

\textminus MoE & 49.88{\fontsize{6.5}{0}\selectfont$\pm$0.4}&72.59{\fontsize{6.5}{0}\selectfont$\pm$0.3}&79.29{\fontsize{6.5}{0}\selectfont$\pm$0.1}&77.56{\fontsize{6.5}{0}\selectfont$\pm$0.9}&84.68{\fontsize{6.5}{0}\selectfont$\pm$0.5}&88.49{\fontsize{6.5}{0}\selectfont$\pm$0.8}&75.44{\fontsize{6.5}{0}\selectfont$\pm$0.4}&87.29{\fontsize{6.5}{0}\selectfont$\pm$0.3}&92.04{\fontsize{6.5}{0}\selectfont$\pm$0.5}&73.87{\fontsize{6.5}{0}\selectfont$\pm$0.9}&83.65{\fontsize{6.5}{0}\selectfont$\pm$0.1}&88.73{\fontsize{6.5}{0}\selectfont$\pm$0.3} \\

\textminus contrast & 49.39{\fontsize{6.5}{0}\selectfont$\pm$0.0}&73.97{\fontsize{6.5}{0}\selectfont$\pm$0.1}&81.71{\fontsize{6.5}{0}\selectfont$\pm$0.4}&74.75{\fontsize{6.5}{0}\selectfont$\pm$0.7}&83.61{\fontsize{6.5}{0}\selectfont$\pm$0.6}&88.55{\fontsize{6.5}{0}\selectfont$\pm$1.3}&72.94{\fontsize{6.5}{0}\selectfont$\pm$0.1}&87.09{\fontsize{6.5}{0}\selectfont$\pm$0.2}&92.90{\fontsize{6.5}{0}\selectfont$\pm$0.2}&73.27{\fontsize{6.5}{0}\selectfont$\pm$0.8}&84.34{\fontsize{6.5}{0}\selectfont$\pm$0.1}&\hl{90.10{\fontsize{6.5}{0}\selectfont$\pm$0.2}} \\

\textminus attention & 49.65{\fontsize{6.5}{0}\selectfont$\pm$0.6}&74.26{\fontsize{6.5}{0}\selectfont$\pm$0.2}&81.54{\fontsize{6.5}{0}\selectfont$\pm$0.3}&78.26{\fontsize{6.5}{0}\selectfont$\pm$1.5}&85.01{\fontsize{6.5}{0}\selectfont$\pm$1.3}&88.60{\fontsize{6.5}{0}\selectfont$\pm$1.1}&76.10{\fontsize{6.5}{0}\selectfont$\pm$0.3}&87.26{\fontsize{6.5}{0}\selectfont$\pm$0.4}&91.53{\fontsize{6.5}{0}\selectfont$\pm$0.5}&74.18{\fontsize{6.5}{0}\selectfont$\pm$0.7}&84.19{\fontsize{6.5}{0}\selectfont$\pm$0.2}&88.86{\fontsize{6.5}{0}\selectfont$\pm$0.7} \\

\textminus mag. norm & 49.25{\fontsize{6.5}{0}\selectfont$\pm$0.9}&73.64{\fontsize{6.5}{0}\selectfont$\pm$0.8}&80.23{\fontsize{6.5}{0}\selectfont$\pm$0.6}&79.56{\fontsize{6.5}{0}\selectfont$\pm$0.4}&86.54{\fontsize{6.5}{0}\selectfont$\pm$0.1}&89.82{\fontsize{6.5}{0}\selectfont$\pm$0.2}&76.13{\fontsize{6.5}{0}\selectfont$\pm$0.4}&\hl{88.40{\fontsize{6.5}{0}\selectfont$\pm$0.3}}&\hl{93.57{\fontsize{6.5}{0}\selectfont$\pm$0.3}}&73.82{\fontsize{6.5}{0}\selectfont$\pm$1.0}&84.17{\fontsize{6.5}{0}\selectfont$\pm$0.8}&89.17{\fontsize{6.5}{0}\selectfont$\pm$1.0} \\

\textminus ind. view & 45.85{\fontsize{6.5}{0}\selectfont$\pm$0.7}&71.78{\fontsize{6.5}{0}\selectfont$\pm$0.3}&80.40{\fontsize{6.5}{0}\selectfont$\pm$0.4}&73.12{\fontsize{6.5}{0}\selectfont$\pm$1.1}&83.15{\fontsize{6.5}{0}\selectfont$\pm$0.3}&88.73{\fontsize{6.5}{0}\selectfont$\pm$0.6}&57.81{\fontsize{6.5}{0}\selectfont$\pm$0.2}&79.49{\fontsize{6.5}{0}\selectfont$\pm$0.3}&88.69{\fontsize{6.5}{0}\selectfont$\pm$0.4}&67.81{\fontsize{6.5}{0}\selectfont$\pm$0.4}&81.98{\fontsize{6.5}{0}\selectfont$\pm$0.4}&87.82{\fontsize{6.5}{0}\selectfont$\pm$0.7} \\

\textminus sep. load & 49.36{\fontsize{6.5}{0}\selectfont$\pm$0.5}&73.82{\fontsize{6.5}{0}\selectfont$\pm$0.8}&81.41{\fontsize{6.5}{0}\selectfont$\pm$0.5}&79.28{\fontsize{6.5}{0}\selectfont$\pm$0.7}&86.29{\fontsize{6.5}{0}\selectfont$\pm$0.8}&90.20{\fontsize{6.5}{0}\selectfont$\pm$0.9}&76.06{\fontsize{6.5}{0}\selectfont$\pm$0.2}&88.14{\fontsize{6.5}{0}\selectfont$\pm$0.2}&93.18{\fontsize{6.5}{0}\selectfont$\pm$0.3}&74.82{\fontsize{6.5}{0}\selectfont$\pm$0.6}&84.26{\fontsize{6.5}{0}\selectfont$\pm$0.3}&88.96{\fontsize{6.5}{0}\selectfont$\pm$0.2} \\

\textminus stop grad & 49.75{\fontsize{6.5}{0}\selectfont$\pm$0.4}&74.10{\fontsize{6.5}{0}\selectfont$\pm$0.1}&81.24{\fontsize{6.5}{0}\selectfont$\pm$0.4}&78.34{\fontsize{6.5}{0}\selectfont$\pm$2.2}&86.10{\fontsize{6.5}{0}\selectfont$\pm$1.0}&89.92{\fontsize{6.5}{0}\selectfont$\pm$1.0}&76.14{\fontsize{6.5}{0}\selectfont$\pm$0.2}&88.20{\fontsize{6.5}{0}\selectfont$\pm$0.3}&93.21{\fontsize{6.5}{0}\selectfont$\pm$0.3}&73.58{\fontsize{6.5}{0}\selectfont$\pm$1.0}&83.07{\fontsize{6.5}{0}\selectfont$\pm$0.4}&88.09{\fontsize{6.5}{0}\selectfont$\pm$0.7} \\

+ full graph & 49.68{\fontsize{6.5}{0}\selectfont$\pm$0.5}&74.21{\fontsize{6.5}{0}\selectfont$\pm$0.5}&81.57{\fontsize{6.5}{0}\selectfont$\pm$0.8}&78.50{\fontsize{6.5}{0}\selectfont$\pm$1.3}&85.27{\fontsize{6.5}{0}\selectfont$\pm$1.5}&88.72{\fontsize{6.5}{0}\selectfont$\pm$1.6}&75.52{\fontsize{6.5}{0}\selectfont$\pm$0.3}&87.43{\fontsize{6.5}{0}\selectfont$\pm$0.0}&92.47{\fontsize{6.5}{0}\selectfont$\pm$0.2}&73.78{\fontsize{6.5}{0}\selectfont$\pm$0.2}&83.88{\fontsize{6.5}{0}\selectfont$\pm$0.2}&88.35{\fontsize{6.5}{0}\selectfont$\pm$0.6} \\

AliAd & \hl{50.29{\fontsize{6.5}{0}\selectfont$\pm$0.4}}&\hl{74.78{\fontsize{6.5}{0}\selectfont$\pm$0.6}}&\hl{81.75{\fontsize{6.5}{0}\selectfont$\pm$0.6}}&\hl{79.68{\fontsize{6.5}{0}\selectfont$\pm$0.3}}&\hl{87.26{\fontsize{6.5}{0}\selectfont$\pm$0.5}}&\hl{91.25{\fontsize{6.5}{0}\selectfont$\pm$0.5}}&\hl{76.33{\fontsize{6.5}{0}\selectfont$\pm$0.1}}&88.18{\fontsize{6.5}{0}\selectfont$\pm$0.2}&93.33{\fontsize{6.5}{0}\selectfont$\pm$0.1}&\hl{74.98{\fontsize{6.5}{0}\selectfont$\pm$0.5}}&\hl{85.31{\fontsize{6.5}{0}\selectfont$\pm$0.5}}&89.72{\fontsize{6.5}{0}\selectfont$\pm$0.2} \\

\Xhline{2\arrayrulewidth}
\end{tabularx}
\caption{Ablation study experimental results. For generality, models are trained on data with missing views.}
\label{tbl: ablation}
\end{table*}
We remove each component to assess its contribution to the whole model. Specifically, we examine the following model variants: (1) without the MoE module, (2) without the contrastive loss, (3) without the attention module, (4) without the magnitude norm, (5) without classification training on individual views, (6) without separate load balancing losses for individual views and the fusion, (7) without the stop gradient function before attention, and (8) with full graph instead of adjusted center contrastive loss.

\Cref{tbl: ablation} presents the results. The poorest results are observed in the variant without individual view training, for which the performance drop is more serious in scenarios with fewer views. The model remains effective when features without magnitude normalization are in a Euclidean space, yielding scores comparable to the proposed version on the RealDisp dataset. Nonetheless, the proposed model still performs better overall compared to this variant.

\subsection{More Results with Random View Missing Rate}
\begin{table*}[]
\fontsize{8}{10}\selectfont
\centering
\begin{tabular}{@{}
lccccccccc
@{}}
\Xhline{2\arrayrulewidth}

Dataset & 
\multicolumn{3}{|c}{DailySport} & 
\multicolumn{3}{|c}{RealDisp} & 
\multicolumn{3}{|c}{UP-Fall} \\

No. views & 
\multicolumn{1}{|c}{1} &
$\leq$ 3 & 
$\leq$ 5 & 
\multicolumn{1}{|c}{1} &
$\leq$ 5 & 
$\leq$ 9 & 
\multicolumn{1}{|c}{1} &
$\leq$ 3 & 
$\leq$ 5 \\

\hline

CMC*                 & 73.66{\fontsize{6.5}{0}\selectfont$\pm$0.8}&80.81{\fontsize{6.5}{0}\selectfont$\pm$0.8}&85.66{\fontsize{6.5}{0}\selectfont$\pm$1.3}&66.41{\fontsize{6.5}{0}\selectfont$\pm$0.8}&80.81{\fontsize{6.5}{0}\selectfont$\pm$0.4}&88.46{\fontsize{6.5}{0}\selectfont$\pm$0.3}&70.25{\fontsize{6.5}{0}\selectfont$\pm$0.4}&77.00{\fontsize{6.5}{0}\selectfont$\pm$0.6}&81.53{\fontsize{6.5}{0}\selectfont$\pm$0.9} \\

COCOA*               & 70.27{\fontsize{6.5}{0}\selectfont$\pm$2.4}&78.50{\fontsize{6.5}{0}\selectfont$\pm$2.6}&84.23{\fontsize{6.5}{0}\selectfont$\pm$2.5}&59.72{\fontsize{6.5}{0}\selectfont$\pm$0.3}&77.52{\fontsize{6.5}{0}\selectfont$\pm$0.1}&86.48{\fontsize{6.5}{0}\selectfont$\pm$0.0}&59.80{\fontsize{6.5}{0}\selectfont$\pm$0.8}&68.16{\fontsize{6.5}{0}\selectfont$\pm$0.6}&73.19{\fontsize{6.5}{0}\selectfont$\pm$0.9} \\

FlexMoE              & 62.42{\fontsize{6.5}{0}\selectfont$\pm$0.1}&75.94{\fontsize{6.5}{0}\selectfont$\pm$0.1}&83.62{\fontsize{6.5}{0}\selectfont$\pm$0.1}&16.71{\fontsize{6.5}{0}\selectfont$\pm$0.1}&48.63{\fontsize{6.5}{0}\selectfont$\pm$0.2}&65.72{\fontsize{6.5}{0}\selectfont$\pm$0.2}&57.54{\fontsize{6.5}{0}\selectfont$\pm$1.8}&75.11{\fontsize{6.5}{0}\selectfont$\pm$1.2}&84.19{\fontsize{6.5}{0}\selectfont$\pm$0.8} \\

FuseMoE              & 60.65{\fontsize{6.5}{0}\selectfont$\pm$4.0}&75.62{\fontsize{6.5}{0}\selectfont$\pm$1.6}&83.97{\fontsize{6.5}{0}\selectfont$\pm$0.7}&27.20{\fontsize{6.5}{0}\selectfont$\pm$1.7}&61.78{\fontsize{6.5}{0}\selectfont$\pm$1.3}&78.56{\fontsize{6.5}{0}\selectfont$\pm$1.3}&56.66{\fontsize{6.5}{0}\selectfont$\pm$0.5}&74.83{\fontsize{6.5}{0}\selectfont$\pm$0.7}&83.97{\fontsize{6.5}{0}\selectfont$\pm$1.0} \\

ShaSpec              & 62.37{\fontsize{6.5}{0}\selectfont$\pm$0.7}&74.17{\fontsize{6.5}{0}\selectfont$\pm$1.1}&81.51{\fontsize{6.5}{0}\selectfont$\pm$2.1}&52.61{\fontsize{6.5}{0}\selectfont$\pm$0.6}&76.07{\fontsize{6.5}{0}\selectfont$\pm$0.1}&87.23{\fontsize{6.5}{0}\selectfont$\pm$0.5}&62.98{\fontsize{6.5}{0}\selectfont$\pm$1.1}&77.25{\fontsize{6.5}{0}\selectfont$\pm$0.6}&85.43{\fontsize{6.5}{0}\selectfont$\pm$0.5} \\

\hline 

\textminus MoE       & 75.47{\fontsize{6.5}{0}\selectfont$\pm$0.9}&82.00{\fontsize{6.5}{0}\selectfont$\pm$0.3}&86.00{\fontsize{6.5}{0}\selectfont$\pm$0.1}&72.39{\fontsize{6.5}{0}\selectfont$\pm$0.0}&82.56{\fontsize{6.5}{0}\selectfont$\pm$0.3}&88.14{\fontsize{6.5}{0}\selectfont$\pm$0.8}&66.09{\fontsize{6.5}{0}\selectfont$\pm$0.4}&72.98{\fontsize{6.5}{0}\selectfont$\pm$0.6}&77.45{\fontsize{6.5}{0}\selectfont$\pm$0.9} \\

\textminus contrast  & 71.64{\fontsize{6.5}{0}\selectfont$\pm$1.1}&79.38{\fontsize{6.5}{0}\selectfont$\pm$0.4}&83.99{\fontsize{6.5}{0}\selectfont$\pm$0.2}&70.30{\fontsize{6.5}{0}\selectfont$\pm$0.6}&83.17{\fontsize{6.5}{0}\selectfont$\pm$0.5}&89.38{\fontsize{6.5}{0}\selectfont$\pm$0.8}&70.34{\fontsize{6.5}{0}\selectfont$\pm$1.1}&79.78{\fontsize{6.5}{0}\selectfont$\pm$0.3}&85.20{\fontsize{6.5}{0}\selectfont$\pm$0.4} \\

\textminus attention & 75.44{\fontsize{6.5}{0}\selectfont$\pm$2.5}&82.42{\fontsize{6.5}{0}\selectfont$\pm$1.9}&86.60{\fontsize{6.5}{0}\selectfont$\pm$1.5}&73.34{\fontsize{6.5}{0}\selectfont$\pm$0.1}&83.32{\fontsize{6.5}{0}\selectfont$\pm$0.4}&88.54{\fontsize{6.5}{0}\selectfont$\pm$0.7}&72.97{\fontsize{6.5}{0}\selectfont$\pm$0.5}&80.07{\fontsize{6.5}{0}\selectfont$\pm$0.0}&85.10{\fontsize{6.5}{0}\selectfont$\pm$0.4} \\

\textminus mag. norm & 73.14{\fontsize{6.5}{0}\selectfont$\pm$0.5}&81.56{\fontsize{6.5}{0}\selectfont$\pm$0.2}&86.90{\fontsize{6.5}{0}\selectfont$\pm$0.1}&\hl{73.94{\fontsize{6.5}{0}\selectfont$\pm$0.3}}&\hl{84.84{\fontsize{6.5}{0}\selectfont$\pm$0.3}}&90.22{\fontsize{6.5}{0}\selectfont$\pm$0.3}&72.69{\fontsize{6.5}{0}\selectfont$\pm$0.3}&80.19{\fontsize{6.5}{0}\selectfont$\pm$0.5}&84.42{\fontsize{6.5}{0}\selectfont$\pm$0.6} \\

\textminus ind. view & 71.92{\fontsize{6.5}{0}\selectfont$\pm$1.3}&81.03{\fontsize{6.5}{0}\selectfont$\pm$1.3}&85.97{\fontsize{6.5}{0}\selectfont$\pm$1.4}&53.89{\fontsize{6.5}{0}\selectfont$\pm$0.1}&74.35{\fontsize{6.5}{0}\selectfont$\pm$0.1}&84.49{\fontsize{6.5}{0}\selectfont$\pm$0.1}&70.47{\fontsize{6.5}{0}\selectfont$\pm$0.6}&78.50{\fontsize{6.5}{0}\selectfont$\pm$0.7}&83.74{\fontsize{6.5}{0}\selectfont$\pm$0.5} \\

\textminus sep. load & 76.36{\fontsize{6.5}{0}\selectfont$\pm$2.1}&83.44{\fontsize{6.5}{0}\selectfont$\pm$1.3}&87.52{\fontsize{6.5}{0}\selectfont$\pm$0.8}&73.52{\fontsize{6.5}{0}\selectfont$\pm$0.2}&84.18{\fontsize{6.5}{0}\selectfont$\pm$0.0}&89.97{\fontsize{6.5}{0}\selectfont$\pm$0.1}&\hl{74.29{\fontsize{6.5}{0}\selectfont$\pm$0.3}}&80.80{\fontsize{6.5}{0}\selectfont$\pm$0.0}&85.06{\fontsize{6.5}{0}\selectfont$\pm$0.2} \\

\textminus stop grad & 75.47{\fontsize{6.5}{0}\selectfont$\pm$1.0}&82.44{\fontsize{6.5}{0}\selectfont$\pm$0.8}&86.75{\fontsize{6.5}{0}\selectfont$\pm$1.7}&72.92{\fontsize{6.5}{0}\selectfont$\pm$0.2}&83.92{\fontsize{6.5}{0}\selectfont$\pm$0.0}&89.45{\fontsize{6.5}{0}\selectfont$\pm$0.0}&73.86{\fontsize{6.5}{0}\selectfont$\pm$0.5}&80.87{\fontsize{6.5}{0}\selectfont$\pm$0.7}&85.56{\fontsize{6.5}{0}\selectfont$\pm$0.0} \\

+full graph          & 73.70{\fontsize{6.5}{0}\selectfont$\pm$0.7}&80.97{\fontsize{6.5}{0}\selectfont$\pm$0.6}&85.97{\fontsize{6.5}{0}\selectfont$\pm$0.8}&73.65{\fontsize{6.5}{0}\selectfont$\pm$0.4}&84.46{\fontsize{6.5}{0}\selectfont$\pm$0.0}&89.91{\fontsize{6.5}{0}\selectfont$\pm$0.1}&72.86{\fontsize{6.5}{0}\selectfont$\pm$0.1}&79.88{\fontsize{6.5}{0}\selectfont$\pm$0.1}&84.03{\fontsize{6.5}{0}\selectfont$\pm$0.0} \\

\hline

AliAd               & \hl{77.40{\fontsize{6.5}{0}\selectfont$\pm$0.6}}&\hl{83.55{\fontsize{6.5}{0}\selectfont$\pm$0.4}}&\hl{87.57{\fontsize{6.5}{0}\selectfont$\pm$0.7}}&73.66{\fontsize{6.5}{0}\selectfont$\pm$0.3}&84.78{\fontsize{6.5}{0}\selectfont$\pm$0.4}&\hl{90.82{\fontsize{6.5}{0}\selectfont$\pm$0.6}}&74.18{\fontsize{6.5}{0}\selectfont$\pm$0.1}&\hl{81.33{\fontsize{6.5}{0}\selectfont$\pm$0.2}}&\hl{86.44{\fontsize{6.5}{0}\selectfont$\pm$0.5}} \\

\Xhline{2\arrayrulewidth}
\end{tabular}
\caption{F1-score (\%) comparison with random data missing rates.}
\label{tbl: exp random missing}
\end{table*}
We conduct another experiment simulating missing views differently. Instead of using a common dropping rate for all views within each dataset, we assign a random dropping rate to each view. Because the missing views of the CMDFall dataset are not simulated, we do not include it in this experiment. Specifically, the dropping rates for views in each dataset are:
\begin{itemize}
    \item Daily Sport: \{"torso": 0.64, "right arm": 0.62, "left arm": 0.77, "right leg": 0.24, "left leg": 0.61\}
    \item RealDisp: \{"right lower arm": 0.5, "right upper arm": 0.54, "back": 0.51, "left upper arm": 0.78, "left lower arm": 0.57, "right calf": 0.54, "right thigh": 0.37, "left thigh": 0.53, "left calf": 0.48\}
    \item UP-Fall: \{"ankle": 0.49, "belt": 0.34, "neck": 0.24, "wrist": 0.63, "2D pose": 0.69\}
\end{itemize}

\Cref{tbl: exp random missing} shows the results for the baseline models, ablation study, and our proposed model. Overall, AliAd achieves the best scores. With a different view missing pattern, these results still remain consistent with the previous experiments.

\section{Other Analyses}
\subsection{Sensitivity Analysis}
\begin{figure}[]
  \centering
  \includegraphics[width=0.93\linewidth]{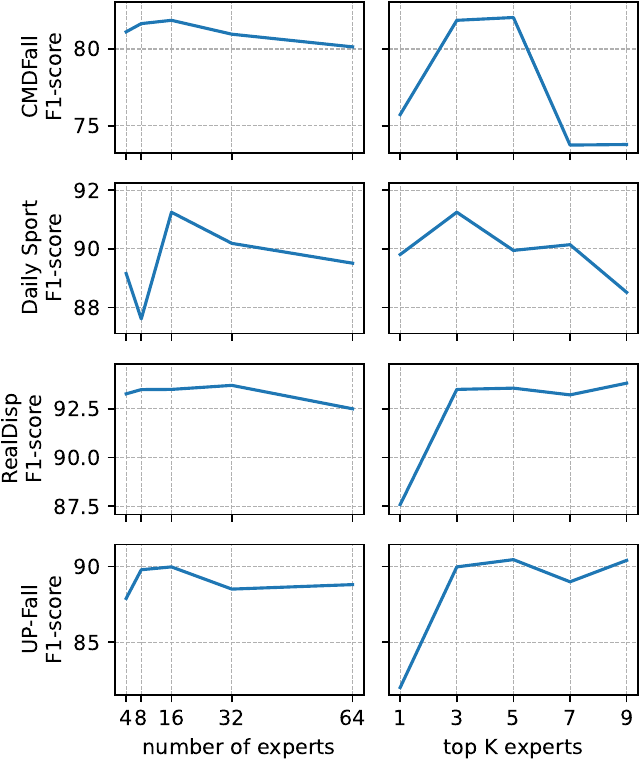}
  \caption{F1-score with varying number of experts and top K experts.}
  \label{fig: num expert sensitivity}
\end{figure}
We perform a sensitivity analysis to evaluate how varying the number of top-K experts and the total number of experts influences model performance. Specifically, we fix the number of top-K experts as specified in \Cref{tbl: hyperparam} while varying the total number of experts, and vice versa. \Cref{fig: num expert sensitivity} shows that using 16 experts yields the best results overall, while increasing the number of experts beyond this point degrades performance in most cases. Regarding top-K selection, the first two datasets experience performance drops as K increases, whereas for the last two datasets, the F1-score remains approximately the same as K increases. Choosing K=3 provides a good balance between accuracy and efficiency.

\subsection{Interaction Between Contrastive Loss and Attention Weights}
\begin{figure*}
  \centering
  \includegraphics[width=\linewidth]{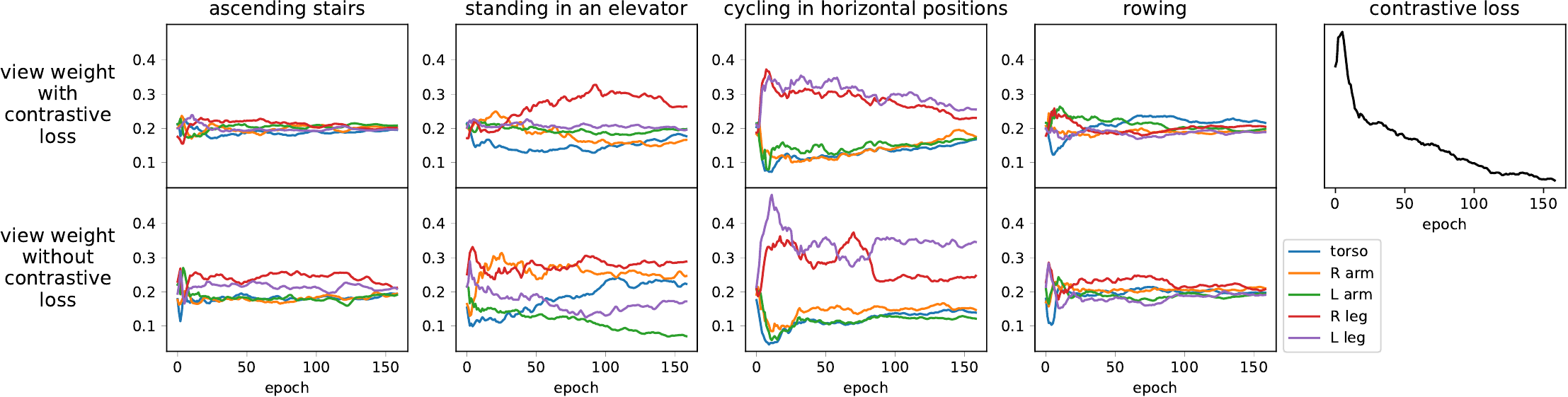}
  \caption{View weights for several classes and contrastive loss over training epochs on the Daily and Sport Activities dataset.}
  \label{fig: view weight change}
\end{figure*}
The attention module looks at angular differences in feature vectors to assign weights to the views. Meanwhile, the contrastive loss aligns all views, making it harder for the attention module to distinguish among them. \Cref{fig: view weight change} illustrates how contrastive loss and view weights evolve over training epochs. As the contrastive loss decreases, the view weights tend to converge toward a similar level. By influencing the upstream encoders, the contrastive loss indirectly impacts the attention module, even though this module is trained exclusively using the classification loss. This behavior confirms that the contrastive loss is functioning as intended, pulling individual views closer to the fusion. In the model without contrastive loss, view weights do not converge or converge more slowly. Models trained with contrastive loss achieve better overall alignment and accuracy (\Cref{tbl: ablation}), highlighting the efficacy of the proposed method.

\subsection{Effects of Separate Load Balancing Loss}
\begin{figure*}[]
  \centering
  \includegraphics[width=0.95\linewidth]{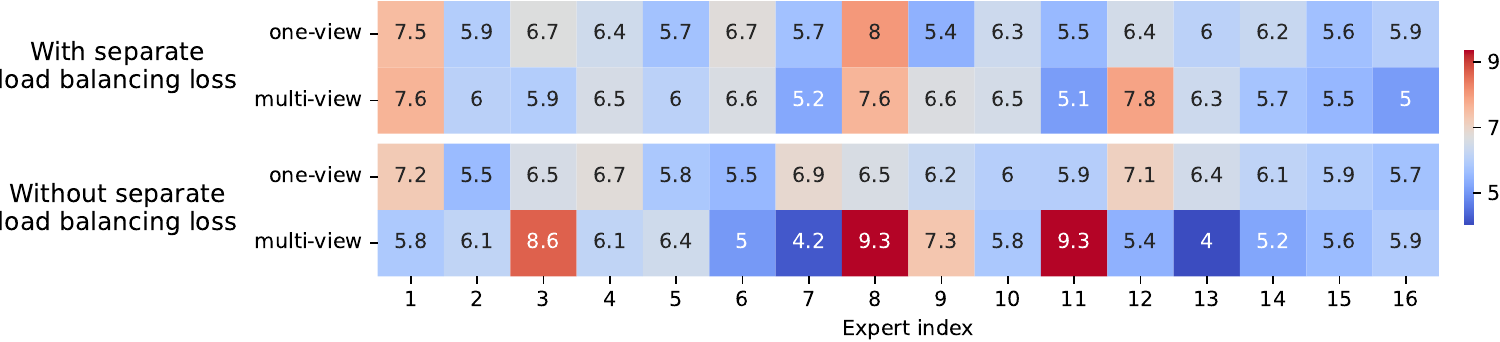}
  \caption{Distribution of gating scores among 16 experts for one-view and multi-view combinations, comparing training with and without separate load balancing loss. Each row is normalized to sum to 100.}
  \label{fig: expert dist}
\end{figure*}
As individual views and the fusion are trained jointly, computing the load balancing loss separately for one-view and multi-view combinations helps distribute tokens more evenly among the experts. This prevents disjoint sets of experts from forming between one-view and multi-view combinations, especially when the contrastive loss is insufficient to align them. To visualize this effect more clearly, we reduced the contrastive loss weight to 0.1 and increased the load balancing loss weight to 1 (from the base hyperparameters in \Cref{tbl: hyperparam}). Each model in this analysis is trained for 10 epochs on the DailySport dataset, with data dropped to simulate missing views. \Cref{fig: expert dist} shows the distribution of gating scores among 16 experts for one-view and multi-view combinations, both with and without the separate load balancing strategy. The gate trained with the separate load balancing loss exhibits more similar distributions between one-view and multi-view combinations, thus achieving the intended effect.

\subsection{Time Complexity Analysis}\label{apx: time complexity}
\begin{figure}[h]
  \centering
  \includegraphics[width=\linewidth]{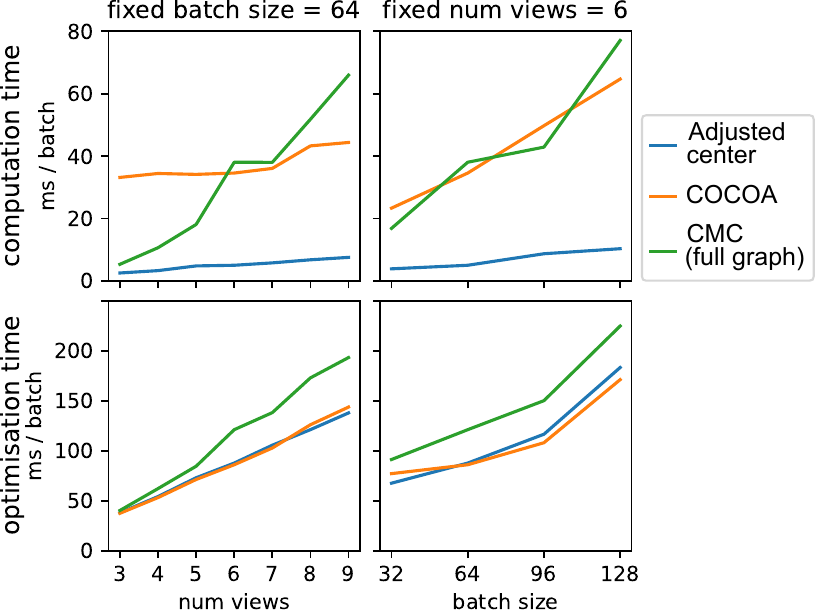}
  \caption{Runtime of contrastive loss methods across varying numbers of views and batch sizes.}
  \label{fig: runtime}
\end{figure}
We empirically compare the time complexity of three contrastive loss functions by measuring the runtime across varying numbers of views and batch sizes, assessing whether observed runtimes align with theoretical complexity. Specifically, we compare the full graph approach\textminus complexity $O(V^2N^2)$ \cite{Tian2020}, the COCOA loss\textminus $O(V^2N{+}VN^2)$ \cite{Deldari2022}, and our adjusted center contrastive loss\textminus $O(VN^2)$. All measurements are conducted on the RealDisp dataset with identical software on the same computer equipped with an Nvidia Quadro RTX4000 GPU and an Intel Xeon Gold 6140 CPU. For each method, we record both the loss computation time and the optimization time per batch, averaging results over 450 consecutive batches. \Cref{fig: runtime} presents the results across a range of view counts and batch sizes. As the number of views increases, the runtimes of COCOA loss and adjusted center contrastive loss scale similarly, while the full graph loss exhibits a substantially steeper increase. With larger batch sizes, the adjusted center loss remains the most efficient, whereas the other two methods show comparable scaling. These empirical results are consistent with theoretical expectations. For optimization time, all methods behave similarly except for the full graph loss, which becomes increasingly costly as the number of views grows.

\section{Conclusion}
This paper introduces AliAd for flexible multimodal multiview HAR. AliAd can handle arbitrarily missing views during both training and inference, while avoiding unnecessary data reconstruction.

The model is trained with a load balancing loss for MoE, an adjusted center contrastive loss, and a classification loss. It leverages both labeled and unlabeled data since contrastive loss does not require labels. Contrastive learning maximizes mutual information among views by aligning different views of the same sample, mitigating the impact of missing views. An attention module assigns view weights, dynamically adjusting view fusion for both contrastive learning and classification. The MoE head addresses residual discrepancies among view combinations that arise from view-specific features and are not captured by contrastive learning.

In HAR, different views often share information as the person's body moves. In rare cases when views share minimal mutual information, contrastive learning's effectiveness would be limited. Our approach samples negative pairs for contrastive loss within each batch, which is effective when the data are diverse. This may be less effective when samples are highly similar, as often seen in modalities like electrocardiography or electroencephalography. These issues need to be addressed when adapting the proposed method to other fields besides HAR.

Our method processes views separately to accommodate missing data. While it retains useful view-specific features, it does not model cross-view interactions, except at the fusion step. Although experiments show it performs favorably compared to methods incorporating cross-view interactions (e.g., using Transformer), this remains an avenue for future exploration, particularly in scenarios with missing views.

\bibliographystyle{unsrt}
\bibliography{main_arxiv}

\end{document}